\documentclass[10pt,twocolumn,letterpaper]{article}

\usepackage{cvpr}              
\definecolor{cvprblue}{rgb}{0.21,0.49,0.74}
\usepackage[pagebackref,breaklinks,colorlinks,allcolors=cvprblue]{hyperref}
\usepackage{colortbl, xcolor,pifont}
\newcommand{\cmark}{\ding{51}} 
\usepackage{textcomp}
\usepackage{multirow}

\def\paperID{***} 
\def\confName{CVPR}
\def\confYear{2026}

\title{RF4D:Neural Radar Fields for Novel View Synthesis in Outdoor Dynamic Scenes}

\author{ Jiarui Zhang, Zhihao Li, Chong Wang, Bihan Wen\thanks{Corresponding author.} \\
  Nanyang Technological University,
  Singapore \\
}

\begin{document}
\maketitle

\begin{abstract}
%
Neural fields (NFs) have achieved remarkable success in scene reconstruction and novel view synthesis.
%
However, existing NF approaches that rely on RGB or LiDAR inputs often struggle under adverse weather conditions, limiting their robustness in real-world outdoor environments such as autonomous driving.
%
In contrast, millimeter-wave radar is inherently resilient to environmental variations, yet its integration with NFs remains largely underexplored.
%
%
Moreover, outdoor driving scenes frequently involve dynamic objects, making spatiotemporal modeling crucial for temporally consistent novel view synthesis.
%
To address these challenges, we present \textbf{RF4D}, a radar-based neural field framework tailored for novel view synthesis in outdoor dynamic scenes.
RF4D explicitly incorporates temporal information into its representation, 
enabling more accurate modeling of object motion.
%
A dedicated \textbf{scene flow module} further predicts motion offsets between adjacent frames, enforcing temporal occupancy coherence during dynamic scene reconstruction.
Moreover, we propose a \textbf{radar-specific power rendering formulation} 
grounded in radar sensing physics, improving both synthesis accuracy and interpretability.
Extensive experiments on public radar datasets demonstrate 
that RF4D substantially outperforms existing methods in radar measurement synthesis and occupancy estimation accuracy, with particularly strong gains in dynamic outdoor environments.
Code is available at \url{https://zhan0618.github.io/RF4D}.

\end{abstract}
\section{Introduction}
\begin{figure}
    \centering
    \includegraphics[width=\linewidth]{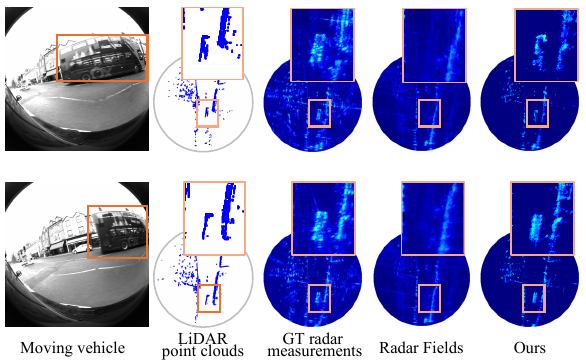}
    \caption{Comparison of radar view synthesis for a dynamic scene with a moving vehicle (orange box). RF4D successfully renders the moving object, whereas Radar Fields~\cite{borts2024radar} fails to recover it.}
    \label{fig:fig1}
\end{figure}
Dynamic scene reconstruction plays a crucial role in numerous applications, including augmented and virtual reality (AR/VR)~\cite{deng2022fov}, robotics~\cite{zhou2023nerf} and autonomous driving~\cite{tonderski2024neurad}.
Recent advances in computer vision~\cite{jiang2025geo4d,wang2024contextgs,li2024chaos, Wang_2025_CVPR,guo2023boundary,guo2023shadowdiffusion,wang2024progressive,fu2026improving,fu2024temporal,tang2025mv} have enabled high-fidelity 3D scene reconstruction and novel view synthesis (NVS), which in turn support downstream tasks such as simulation~\cite{yang2023reconstructing}, motion planning~\cite{hu2022st,hu2023planning}, and scene understanding~\cite{sadat2020perceive,muhammad2022vision}.
%
For instance, dynamic scene reconstruction enables closed-loop simulations for training and evaluating end-to-end planning algorithms under diverse, real-world conditions~\cite{zhao2024drive}.
%
While NVS has been widely studied using camera~\cite{ost2021neural,turki2023suds,wang2023neural} and LiDAR data~\cite{tao2024lidar,zheng2024lidar4d}, radar sensing remains relatively underexplored despite its extensive use in autonomous driving.
%
Radar offers unique advantages, such as robustness to adverse weather and low light, long-range perception, and cost efficiency~\cite{venon2022millimeter,fan2024diffusion}.
%
Similar to LiDAR and RGB sensors, radar also provides partial and viewpoint-dependent observations, making it a natural candidate for scene reconstruction.
%
%
However, radar data is particularly challenging due to its low spatial resolution, sparsity, internal noise, and multipath effects~\cite{richards2005fundamentals,fan2025mmpred,fan2025m4human}.
%
The presence of moving objects in outdoor driving scenes further complicates the reconstruction process.

%
%
Traditional radar-based reconstruction methods typically convert individual radar measurements into sparse point clouds, aggregate them across frames, and align them in a global coordinate system.
%
To enable novel view synthesis, prior works often rely on simulation-based techniques such as ray tracing or radio frequency (RF) propagation modeling~\cite{arnold2022maxray, schussler2021realistic}.
%
However, these approaches rely heavily on hardware-specific radar parameters and waveform configurations, limiting their generalizability~\cite{bialer2024radsimreal}.
%
Moreover, they are mostly constrained to static scenes and struggle to capture view-dependent radar cross section (RCS) variations observed in real-world dynamic environments.

\begin{figure}
    \centering
    \includegraphics[width=.9\linewidth]{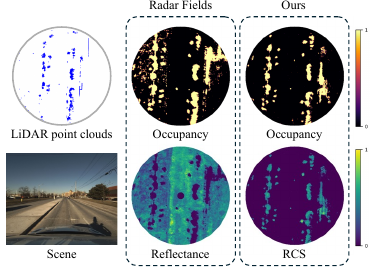}
    \caption{Predicted occupancy and reflectance from Radar Fields~\cite{borts2024radar} versus occupancy and radar cross-section (RCS) from RF4D. Our predictions follow radar physics, where high occupancy corresponds to strong RCS, while Radar Fields lacks such consistency.}
    \label{fig:fig1b}
\end{figure}

%
%
%
Neural fields (NFs) have recently emerged as a powerful paradigm for implicit scene representation, enabling breakthroughs in both reconstruction and NVS.
%
The seminal Neural Radiance Fields (NeRF)~\cite{mildenhall2021nerf} integrates NFs with optical volume rendering, achieving remarkable results for RGB-based reconstruction.
%
Extending this idea, Radar Fields~\cite{borts2024radar} introduced a radar-based NF model that replaces optical rendering with a physics-driven forward model to predict received radar power.
However, they are limited to \textbf{static scenes}, leading to failure cases in dynamic environments (\eg, disappearing moving vehicles, as shown in Figure~\ref{fig:fig1}).
%
Furthermore, we observe a fundamental issue in their design, \ie, an \textbf{occupancy–reflectance contradiction}, where regions with high predicted occupancy often exhibit suppressed reflectance, as illustrated in Figure~\ref{fig:fig1b}.
%
This behavior violates the physical expectation that occupied regions should produce stronger reflections.
%
%
These models also rely on \textbf{external occupancy estimators} for supervision, introducing additional dependencies.

%
To address these limitations, we propose \textbf{RF4D}, a radar-based neural field framework designed for NVS in outdoor dynamic scenes.
%
Specifically, RF4D represents each scene as a spatiotemporal NF parameterized by both position and time, and predicts two radar-specific properties at each spatial point: (1) \textbf{occupancy}, indicating whether the point is occupied, and (2) \textbf{RCS}, describing the reflectivity of the occupied region.
%
To ensure temporal consistency, we introduce a \textbf{scene flow module} that predicts motion offsets, promoting stable and coherent occupancy predictions across adjacent frames.
%
To resolve the occupancy–reflectance contradiction, we further propose a \textbf{radar-specific power rendering formulation} grounded in radar sensing physics, which estimates the received radar power based on the predicted occupancy and RCS.
%
This formulation maintains a physically faithful relationship between occupancy and reflectance, while eliminating the need for external occupancy supervision.
%
As illustrated in Figure~\ref{fig:fig1b}, RF4D produces predictions consistent with radar physics, \ie, high occupancy corresponds to strong reflectance, unlike the contradictory behavior seen in prior work.

In summary, our contributions are three fold: 
\begin{itemize}
    \item We propose \textbf{RF4D}, a radar-based neural field framework for novel view synthesis in dynamic outdoor scenes, representing the scene as a continuous function of both spatial and temporal dimensions. A scene flow module promotes temporal coherence by learning motion offsets across frames.
    
    \item We design a \textbf{radar-specific power rendering formulation} that jointly leverages predicted occupancy and RCS, preserving physically consistent relationships and removing the dependence on external occupancy supervision.
    
    \item We conduct extensive experiments on two public radar datasets, demonstrating that RF4D achieves state-of-the-art performance in radar measurement synthesis and occupancy estimation, particularly in dynamic scenarios.
\end{itemize}

\begin{figure*}
    \centering
    \includegraphics[width=.95\linewidth]{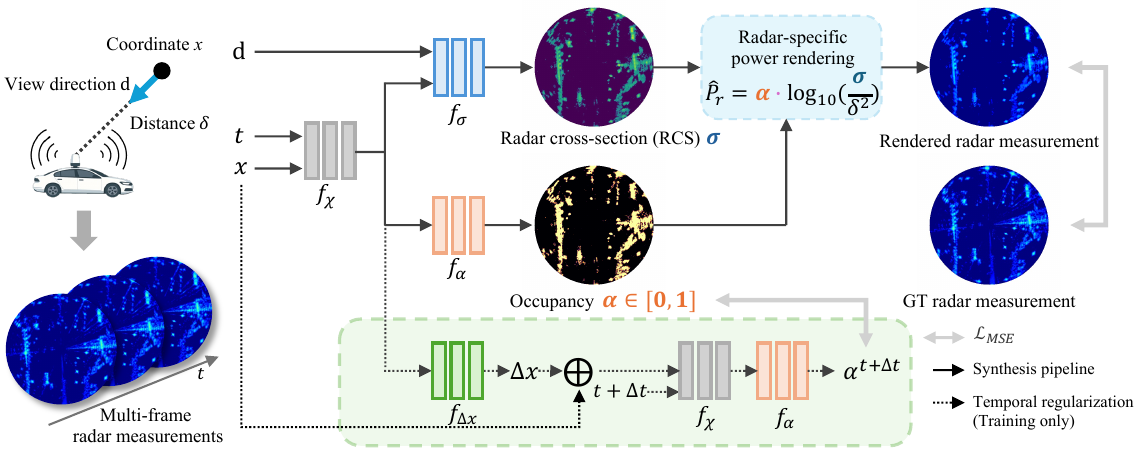}
    \caption{Overview of the proposed RF4D framework. Given a 3D query point $x$ at time $t$ and view direction $\mathbf{d}$, RF4D first predicts two radar-specific physical quantities: occupancy $\alpha$ and radar cross-section (RCS) $\sigma$, using neural radar fields. The occupancy $\alpha$ indicates whether the point is physically occupied, and the RCS $\sigma$ represents its reflectivity. These quantities are combined through the radar-specific power rendering to estimate the received radar power. During training, the rendered power is supervised by ground-truth radar measurements, and the scene flow module enforces temporal consistency by predicting motion offsets and warping points to adjacent frames to regularize occupancy over time.
 }
    \label{fig:fig2}
\end{figure*}

\section{Related works}
\subsection{Radar simulation}
Radar simulation aims to generate realistic radar measurements for scenes. Traditional radar simulation methods rely on physical modeling of radar sensors and the environments through ray tracing and Radio Frequency propagation modeling \cite{arnold2022maxray,dudek2010millimeter,hirsenkorn2017ray,holder2019fourier,schoffmann2021virtual,schussler2021realistic,thieling2020scalable}, but they require detailed knowledge of radar hardware and do not scale across sensors. Generative models, such as generative adversarial networks (GAN) or variational autoencoders (VAE), 
have been used to learn radar data distributions from real measurements, but they depend on large annotated datasets and sensor-specific tuning \cite{weston2021there,wang2020l2r,wheeler2017deep}. RadSimReal \cite{bialer2024radsimreal} introduces a learning-based radar simulator that synthesizes realistic radar signals from 3D ground-truth scenes, without requiring detailed radar hardware modeling.
Recent efforts have adapted neural fields to radar by modeling occupancy and reflectance. NeuRadar~\cite{rafidashti2025neuradar} jointly models camera, LiDAR, and radar, and uses camera/LiDAR cues to guide radar point cloud rendering.
DART~\cite{huang2024dart} represents range-Doppler measurements using neural fields collected with handheld radar. Radar Fields~\cite{borts2024radar} targets range-azimuth representations using ego-motion radar, relying on external occupancy estimators~\cite{werber2015automotive} for regularization. RadarSplat~\cite{kung2025radarsplat} advances detailed physical modeling of radar to synthesize realistic measurements, including multipath and noise. This direction is orthogonal to our approach and is broadly beneficial to radar NVS. Overall, existing methods address static scenes only and do not model dynamics.

\subsection{Radar occupancy estimation}
Radar occupancy estimation initially relies on CFAR-based methods~\cite{richards2005fundamentals}, which detect local peaks in radar range-azimuth or range-Doppler maps and project them into Cartesian space. However, CFAR is highly sensitive to noise and multipath reflections, resulting in sparse and unreliable maps. To address this, Bayesian filtering approaches were introduced, modeling occupancy as a probabilistic state and enabling temporal fusion to account for uncertainty and partial observability \cite{werber2015automotive}. More recently, data-driven methods have emerged~\cite{brodeski2019deep,cao2021dnn,kung2022radar,sless2019road,weston2019probably,engelhardt2019occupancy,ding2024radarocc}, but they typically require ground-truth occupancy maps derived from LiDAR for supervision during training. In contrast, our method estimates occupancy solely from radar measurements, eliminating the need for generating labels.
\subsection{Dynamic scene reconstruction using NFs}
Most existing works on NVS in dynamic scenes focus on RGB images as input modalities. These approaches can be broadly categorized into two types. The first type models scene motion through deformation fields, which warp observations at each time step into the predefined canonical space, enabling consistent geometry and appearance modeling across time~\cite{park2021nerfies,park2021hypernerf,pumarola2021d}. They use learned deformation networks to handle non-rigid motion and occlusions. The second type takes time as an explicit input, learning a direct mapping from spatiotemporal coordinates to dynamic geometry and radiance, such as~\cite{li2021neural,fridovich2023k,shao2023tensor4d,li2023dynibar,li2022neural}, offering a more implicit and continuous representation of dynamic scenes.
In contrast, radar-based NVS in dynamic scenes remains largely underexplored. To the best of our knowledge, our work is the first to model dynamic scenes from radar measurements using neural fields.

\section{Preliminary}

\paragraph{Radar measurements and physical model.}
Our work uses radar data collected by a mechanically spinning Frequency-Modulated Continuous Wave (FMCW) radar operating at millimeter-wave (mmWave) frequencies. 
The radar performs a full 360° horizontal sweep, capturing returns over a set of azimuth directions during each rotation.
Each radar measurement is represented as a 2D range-azimuth map with shape $N_\theta \times N_\delta$, where $N_\theta$ denotes the number of discrete azimuth directions (beams), and $N_\delta$ denotes the number of range bins along each beam. 
The resulting map is a polar image, where each pixel corresponds to the received signal power at a specific azimuth direction and range.
The value at a given bin reflects the received power from objects located at that range and direction. 
The received power $P_r$ from a target at range $\delta$ is given by:
\begin{equation}\label{eq:phsyics_model}
P_r = \frac{P_t \cdot G^2 \cdot \sigma}{(4\pi)^3 \delta^2},
\end{equation}
where  
$P_t$ is the transmitted power,  
$G$ is the antenna gain,  
$\delta$ is the distance to the target, and  
$\sigma$ is the radar cross section (RCS) of the target.
The RCS $\sigma$ quantifies the amount of incident radar energy reflected back toward the receiver, in short, it measures the target's reflectivity.
It depends on the target's physical characteristics, including size, shape, material, and orientation relative to the incoming signal. 
The inverse square power dependence on $\delta$ reflects the signal attenuation over distance.

\begin{table*}[t]
\centering
\scriptsize
\setlength{\tabcolsep}{2pt}
\caption{Quantitative comparison across four different driving scenarios from the RobotCar dataset~\cite{RadarRobotCarDatasetICRA2020}.
The best result and the runner-up are highlighted in \textbf{bold} and \underline{underline}, respectively.}
\begin{tabular}{c|cccc|cccc|cccc|cccc}
\toprule
& \multicolumn{4}{c|}{\textbf{Scene 1}} 
& \multicolumn{4}{c|}{\textbf{Scene 2}} 
& \multicolumn{4}{c}{\textbf{Scene 3}}
& \multicolumn{4}{c}{\textbf{Scene 4}} \\
\textbf{Method}  & SSIM\textuparrow & PSNR\textuparrow & CD\textdownarrow & RCD\textdownarrow
                 & SSIM\textuparrow & PSNR\textuparrow & CD\textdownarrow & RCD\textdownarrow
                 & SSIM\textuparrow & PSNR\textuparrow & CD\textdownarrow & RCD\textdownarrow
                 & SSIM\textuparrow & PSNR\textuparrow & CD\textdownarrow & RCD\textdownarrow\\
\midrule
D-NeRF~\cite{pumarola2021d}  
& 0.1270 & 20.80 & 80.9668 & 0.1046
& 0.1363 & 20.29 & 96.7480 & 0.1114
& 0.1620 & 21.27 & 23.8396 & 0.1198
& 0.0867 & 17.58 & 69.4628 & 0.1710 \\

Hexplane~\cite{cao2023hexplane}   
& 0.2674 & 21.12 & 78.9851 & 0.0879
& 0.2980 & 20.21 & 84.3646 & 0.2238
& \underline{0.3909} & 22.09 & 119.0078 & 0.2068
& 0.2267 & 20.25 & 58.8482 & 0.2207 \\
\midrule
RadarFields~\cite{borts2024radar} 
& \underline{0.3372} & \underline{23.36} & \underline{18.1915} & \underline{0.0300}
& \underline{0.2997} & \textbf{22.24} & \underline{16.1489} & \underline{0.0149}
& 0.3498 & \textbf{24.02} & \underline{9.5357} & \underline{0.0281}
& 0.2829 & \underline{21.95} & \underline{20.9821} & \underline{0.0219}\\

Ours       
& \textbf{0.6103} & \textbf{23.38} & \textbf{7.3449} & \textbf{0.0190}
& \textbf{0.5350} & \underline{22.16} & \textbf{8.0909} & \textbf{0.0125}
& \textbf{0.6258} & \underline{23.76} & \textbf{3.2896} & \textbf{0.0097}
& \textbf{0.4971} & \textbf{22.20} & \textbf{18.7607} & \textbf{0.0147} \\
\bottomrule
\end{tabular}
\label{tab:oxford_results}
\end{table*}

\begin{table*}[t]
\centering
\scriptsize
\setlength{\tabcolsep}{2pt}
\caption{Quantitative comparison across four driving scenarios from the Boreas dataset~\cite{burnett_ijrr23}. The best result and the runner-up are highlighted in \textbf{bold} and \underline{underline}, respectively. CD and RCD are not reported for the snow scene due to \textit{unreliable ground-truth LiDAR geometry under heavy snowfall}.}
\begin{tabular}{c|cccc|cccc|cccc|cccc}
\toprule
 & \multicolumn{4}{c|}{\textbf{Sun}} 
  & \multicolumn{4}{c|}{\textbf{Snow}} 
  & \multicolumn{4}{c|}{\textbf{Rain}}
  & \multicolumn{4}{c}{\textbf{Static}} \\
\textbf{Method}  & SSIM\textuparrow & PSNR\textuparrow & CD\textdownarrow & RCD\textdownarrow 
               & SSIM\textuparrow & PSNR\textuparrow & CD\textdownarrow & RCD\textdownarrow 
               & SSIM\textuparrow & PSNR\textuparrow & CD\textdownarrow & RCD\textdownarrow 
               & SSIM\textuparrow & PSNR\textuparrow & CD\textdownarrow & RCD\textdownarrow \\
\midrule
D\mbox{-}NeRF~\cite{pumarola2021d}

 & 0.0923 & 12.33 & 144.7542 & 1.4754
 & 0.1676 & 20.48 & - & -
 & 0.0991 & 13.31 & 435.1305 & 1.1928
 & 0.1483 & 19.65 & 46.1635 & 0.2127\\

Hexplane~\cite{cao2023hexplane}

 & 0.2892 & 22.90 & 28.2160 & 0.0911
 & 0.2643 & 23.61 & - & -
 & 0.2946 & 23.28 & 38.5217 & 0.2149
 & 0.2388 & 24.23 & 46.1504 & 0.2114\\
\midrule
RadarFields~\cite{borts2024radar}

 & \underline{0.3641} & \underline{25.39} & \underline{10.6385} & \underline{0.0060}
 & \underline{0.4823} & \underline{27.36} & - & -
 & \underline{0.3905} & \underline{25.80} & \underline{23.8941} & \underline{0.0312} 
 & \underline{0.4331} & \underline{26.96} & \underline{14.1116} & \underline{0.0069}\\

Ours
 
& \textbf{0.7001} & \textbf{26.65} & \textbf{5.4482} & \textbf{0.0033} 
& \textbf{0.7946} & \textbf{27.98} & - & - 
& \textbf{0.6635} & \textbf{26.21} & \textbf{10.4558} & \textbf{0.0051} 
& \textbf{0.7724} & \textbf{27.55} & \textbf{8.0739} & \textbf{0.0040} \\
\bottomrule
\end{tabular}
\label{tab:boreas_results}
\end{table*}

\begin{table*}[t]
\centering
\scriptsize
\setlength{\tabcolsep}{2pt}
\caption{Ablation study for the RobotCar dataset~\cite{RadarRobotCarDatasetICRA2020}. 
The best result and the runner-up are highlighted in \textbf{bold} and \underline{underline}, respectively.}

\begin{tabular}{ccc|cccc|cccc|cccc|cccc}
\toprule
  \multicolumn{3}{c|}{~} 
& \multicolumn{4}{c|}{\textbf{Scene 1}} 
& \multicolumn{4}{c|}{\textbf{Scene 2}} 
& \multicolumn{4}{c}{\textbf{Scene 3}} 
& \multicolumn{4}{c}{\textbf{Scene 4}} \\
$\mathcal{L}_{p}$ & $\mathcal{L}_{oc}$ & $\mathcal{L}_m$ 
& PSNR\textuparrow & SSIM\textuparrow & CD\textdownarrow & RCD\textdownarrow 
& PSNR\textuparrow & SSIM\textuparrow & CD\textdownarrow & RCD\textdownarrow 
& PSNR\textuparrow & SSIM\textuparrow & CD\textdownarrow & RCD\textdownarrow  
& PSNR\textuparrow & SSIM\textuparrow & CD\textdownarrow & RCD\textdownarrow\\
\midrule
& & 
& 21.90 & 0.3184 & 80.9668 & 0.1046 
& 20.35 & 0.2616 & 96.7480 & 0.1114 
& \textbf{23.80} & 0.4161 & 23.8396 & 0.1198 
& \textbf{22.55} & 0.3586 & 45.3407 & 0.1174 \\

\cmark & &  
& \underline{23.23} & 0.5662 & 9.2251 & 0.0200 
& \underline{22.10} & \underline{0.5315} & 11.8233 & 0.0145
& 23.56 & \underline{0.6240} & 6.5126 & 0.0108 
& 22.10 & \textbf{0.5152} & 23.8087 & 0.0176 \\

\cmark & \cmark &   
& 22.99 & \textbf{0.6167} & \underline{7.5713} & \textbf{0.0150} 
& 22.08 & 0.5269 & \underline{10.4477} & \underline{0.0137}
& 23.33 & 0.6182 & \underline{5.1028} & \underline{0.0098} 
& 21.58 & 0.4933 & \underline{19.5720} & \underline{0.0150} \\

\cmark & \cmark & \cmark 
& \textbf{23.38} & \underline{0.6103} & \textbf{7.3449} & \underline{0.0190} 
& \textbf{22.16} & \textbf{0.5350} & \textbf{8.0909} & \textbf{0.0125} 
& \underline{23.76} & \textbf{0.6258} & \textbf{3.2896} & \textbf{0.0097} 
& \underline{22.20} & \underline{0.4971} & \textbf{18.7607} & \textbf{0.0147} \\
\bottomrule
\end{tabular}
\label{tab:oxford_ablation}
\end{table*}

\begin{table*}[t]
\centering
\scriptsize
\setlength{\tabcolsep}{2pt}
\caption{Ablation study for the Boreas dataset~\cite{burnett_ijrr23}. The best result and the runner-up are highlighted in \textbf{bold} and \underline{underline}, respectively. CD and RCD are not reported for the snow scene due to \textit{unreliable ground-truth LiDAR geometry under heavy snowfall}.}

\begin{tabular}{ccc|cccc|cccc|cccc|cccc}
\toprule
  \multicolumn{3}{c|}{~} 
& \multicolumn{4}{c|}{\textbf{Sun}} 
& \multicolumn{4}{c|}{\textbf{Snow}} 
& \multicolumn{4}{c}{\textbf{Rain}} 
& \multicolumn{4}{c}{\textbf{Static}} \\
$\mathcal{L}_{p}$ & $\mathcal{L}_{oc}$ & $\mathcal{L}_m$ & PSNR \textuparrow & SSIM \textuparrow & CD\textdownarrow & RCD\textdownarrow & PSNR \textuparrow& SSIM\textuparrow & CD\textdownarrow & RCD\textdownarrow & PSNR\textuparrow & SSIM\textuparrow & CD\textdownarrow & RCD\textdownarrow  &PSNR\textuparrow & SSIM\textuparrow & CD\textdownarrow & RCD\textdownarrow\\
\midrule
& & 
& 25.12 & 0.5662 & 28.6160 & 0.0911 
& 27.16 & 0.5967 & - & -
& 25.50 & 0.5528 & 38.5127 & 0.2149
& \underline{27.07} & 0.6417 & 46.1504 & 0.2144 \\

\cmark & & 
& \underline{25.28} & \underline{0.6910} & 6.3486 & 0.0048
& \underline{27.52} & \underline{0.7907} & - & - 
& \underline{25.75}& \underline{0.6588} & 16.8615 & 0.0102 
& 26.58 & \underline{0.7488} & \underline{8.9845} & \underline{0.0058} \\

\cmark & \cmark &  & 
23.96 & 0.6525 & \underline{5.9112} & \underline{0.0040} 
& 23.43 & 0.7094 & - & -
& 23.46 & 0.6009 & \underline{15.0229} & \underline{0.0099}
& 23.68 & 0.6510 & 13.2131 & 0.0185\\

\cmark & \cmark & \cmark 
& \textbf{26.65} & \textbf{0.7001} & \textbf{5.4482} & \textbf{0.0033} & \textbf{27.98} & \textbf{0.7946} & - & - 
& \textbf{26.21} & \textbf{0.6635} & \textbf{10.4558} & \textbf{0.0051} 
& \textbf{27.55} & \textbf{0.7724} & \textbf{8.0739}& \textbf{0.0040} \\
\bottomrule
\end{tabular}
\label{tab:boreas_ablation}
\end{table*}

\section{Method}
%

\subsection{Problem Formulation}\label{problem_formulation}
In dynamic driving scenarios, we are given a sequence of radar measurements  
$S = \{S_0, S_1, ..., S_{n-1}\}$, where $S_i \in \mathbb{R}^{N_\theta \times N_\delta}$,  
along with the corresponding radar poses  
$H_s = \{H_0, H_1, ..., H_{n-1}\}$, where $H_i \in \mathrm{SE}(3)$,  
and timestamps  
$T_s = \{t_0, t_1, ..., t_{n-1}\}$, where $t_i \in \mathbb{R}$.
Here, $N_\theta$ denotes the number of azimuth directions (beams), and $N_\delta$ is the number of range bins per beam. 
Each bin records a 1D reflected power value measured at the corresponding azimuth and range.

The goal of RF4D is to reconstruct this dynamic scene as a continuous implicit representation based on neural fields.
Furthermore, given a novel radar pose $H_{novel}$ and any moment $t_{novel}$, RF4D performs neural rendering to synthesize the radar measurement $S_{novel}$.
\subsection{RF4D}\label{RF4D}
The overall framework of RF4D is illustrated in Figure~\ref{fig:fig2}.
\paragraph{Radar-to-world projection.}
Each radar measurement $S_i \in \mathbb{R}^{N_\theta \times N_\delta}$ is a 2D range-azimuth map, where each bin indexed by $(\theta_j, \delta_k)$ corresponds to a beam direction $\theta_j$ and range bin $\delta_k$. 
To interpret radar measurements in 3D space, we first convert each bin into a Cartesian coordinate in the radar’s local frame:
$
\mathbf{x}_{\text{radar}} = [
\delta_k  \cos(\theta_j)~
\delta_k  \sin(\theta_j)~
0]^T.
$
The point is then transformed into the global world coordinate frame using the radar pose $H_i \in \mathrm{SE}(3)$ associated with measurement $S_i$ such that 
$
\mathbf{x}_{\text{world}} = H_i \cdot \mathbf{x}_{\text{radar}}
$.
In addition to the 3D location $\mathbf{x}_{\text{world}}$, we also compute the view direction $\mathbf{d}$ for each point, which is the normalized vector from the radar origin to that point.
%
This process enables us to associate each range-azimuth bin with a world-coordinate 3D position and its corresponding direction, which are used as inputs to our neural field for rendering and training.

\paragraph{Neural radar fields.}
The input consists of a timestamp $t$, a 3D spatial location $\mathbf{x}=(x,y,z)$ and a view direction $\mathbf{d}$.
The output includes: occupancy $\alpha$, indicating whether the position is occupied or free, and radar cross section (RCS) $\sigma$, which quantifies the visibility of a target to the radar.

We encode the position $\mathbf{x}$ using a multi-resolution hash grid ~$\mathcal{H}$~\cite{muller2022instant}, and the timestamp $t$ with a learnable embedding network $\mathcal{T}$.
The encodings are then concatenated and passed to an MLP $f_\chi$ to obtain a spatialtemporal latent feature $\chi$:
\begin{equation}
    \chi = f_\chi(\mathcal{H}(\mathbf{x}), \mathcal{T}(t)).
\end{equation}
The latent feature $\chi$ is then passed to $f_\alpha$, an MLP that predicts occupancy using a Gumbel-Sigmoid activation~\cite{jang2016categorical} at the output.
Note that Gumbel-Sigmoid is to encourage the predicted $\alpha$ to approximate binary values, functioning as a soft binary mask aligned with physical interpretation of occupancy: 0 implies empty and 1 implies occupied.
Since RCS also depends on the incident angle of radar waveform (view direction), to predict RCS $\sigma$, we concatenate $\chi$ with the encoded direction $\mathbf{d}$ and pass them to $f_\sigma$. Here we use spherical harmonics $\mathcal{S}$~\cite{yu2021plenoctrees} to encode $\mathbf{d}$.
Finally, we obtain the two predicted radar-specific characteristics:
\begin{equation}
\begin{aligned}
\text{Occupancy}~\alpha = f_\alpha(\chi), \\
\text{RCS}~\sigma = f_\sigma(\chi,\mathcal{S}(\mathbf{d})).
\end{aligned}
\end{equation}
\paragraph{Temporal regularization.}
To ensure temporal consistency in occupancy predictions across frames, we introduce a scene flow module that models the motion of scene points over adjacent frames.
Given a latent feature representation $\chi$ at time $t$, this module predicts motion offsets $\Delta x$ through an MLP $f_{\Delta x}$:
\begin{equation}
    \Delta x = f_{\Delta x}(\chi).
\end{equation}
The offset $\Delta x \in \mathbb{R}^6$ consists of two parts: the first three dimensions represent the offset to the previous frame $\Delta x^{-}$, and the last three dimensions represent the offset to the next frame $\Delta x^{+}$.
Each spatial point is then warped to its estimated position in the adjacent frames.
The warped points are passed through the feature MLP $f_\chi$ and the occupancy MLP $f_\alpha$ to estimate the occupancies at adjacent frames:
\begin{equation}\label{eq:alpha_neighbour}
    \begin{aligned}
        \alpha^{t - \Delta t} = f_\alpha(f_{\chi}(\mathcal{H}(x+\Delta x^-),\mathcal{T}( t-\Delta t))), \\
        \alpha^{t + \Delta t} = f_\alpha(f_{\chi}(\mathcal{H}(x+\Delta x^+), \mathcal{T}(t+\Delta t))).
    \end{aligned}
\end{equation}
We regularize these occupancy predictions across time to promote temporal coherence. 
This encourages the model to produce stable and consistent occupancy, especially for moving objects.

\paragraph{Radar-specific power rendering.}
Synthesizing radar signals requires incorporating the radar sensing physics and cannot directly apply traditional optical volume rendering in NeRF.
The received radar power follows the model in Equation~\ref{eq:phsyics_model}, where power depends on the radar cross-section (RCS) $\sigma$ of the target and its distance $\delta$ from the sensor.
Following \cite{borts2024radar}, constant factors such as transmitted power and antenna gain are omitted since they remain fixed during training.
A base-10 logarithmic transformation is applied to match the decibel scale commonly used in radar data.
Instead of disentangling occupancy and reflectance like ~\cite{borts2024radar}, we treat the predicted occupancy $\alpha$ as a soft gating mask that activates radar responses only when a target is predicted physically present.
This mechanism suppresses noise and reduces the effect of multipath interference.
The resulting rendering is given by:
\begin{equation}
\hat{P}_r = \alpha \cdot \log_{10}\left(\frac{\sigma}{\delta^2}\right).
\end{equation}
\subsection{Training objective}\label{sec:training_objective}
The training objective combines four components: a radar power reconstruction loss, a temporal consistency regularization, an occupancy sparsity regularization, and a motion offset regularization.
During each iteration, $N$ range-azimuth bins $(\delta, \theta)$ are sampled.
Each range-azimuth bin receives predicted occupancy $\hat{\alpha}_{\delta,\theta}$, RCS $\hat{\sigma}_{\delta,\theta}$, and motion offsets $\Delta x_{\delta,\theta}^-$ and $\Delta x_{\delta,\theta}^+$. The rendered power is
$\hat{P}_{\delta,\theta} = \hat{\alpha}_{\delta,\theta} \cdot \log_{10}(\hat{\sigma}_{\delta,\theta} / \delta^2)$.

To supervise radar signal synthesis, we minimize the mean squared error (MSE) loss between the rendered power and the ground-truth measurement $P_{GT_{\delta,\theta}}$ over all the sampled range-azimuth bins:
\begin{equation}
    \mathcal{L}_{\text{rt}} = \frac{1}{N} \sum_{\delta,\theta} 
    \left( \hat{\alpha}_{\delta,\theta} \cdot \log_{10}\left( \frac{\hat{\sigma}_{\delta,\theta}}{\delta^2} \right)
    - P_{GT_{\delta,\theta}} \right)^2.
\end{equation}

To encourage temporal coherence, the predicted occupancy at time $t$ is encouraged to remain consistent with its counterparts in the adjacent frames.
This is achieved through a temporal regularization term:
\begin{equation}
    \mathcal{L}_{\text{oc}} = \frac{1}{N} \sum_{\delta,\theta} \left(
    \left( \hat{\alpha}_{\delta,\theta} - \hat{\alpha}^{t - \Delta t}_{\delta,\theta} \right)^2 + 
    \left( \hat{\alpha}_{\delta,\theta} - \hat{\alpha}^{t + \Delta t}_{\delta,\theta} \right)^2 \right),
\end{equation}
where $\hat{\alpha}^{t - \Delta t}_{\delta,\theta}$ and $\hat{\alpha}^{t + \Delta t}_{\delta,\theta}$ are the predicted occupancies of the same point at adjacent frames obtained through Equation~\ref{eq:alpha_neighbour}.

To avoid degenerate predictions (e.g., $\hat{\alpha}_{\delta,\theta} \approx 1$ everywhere), we impose a sparsity regularization by penalizing the mean occupancy over the sampled range-azimuth bins:
\begin{equation}
    \mathcal{L}_{\text{p}} =  \frac{1}{N} 
    \sum_{\delta,\theta} \hat{\alpha}_{\delta,\theta}.
\end{equation}

Finally, the predicted motion offsets are regularized to prevent unrealistic displacements.
We apply a regularization on the magnitude of offsets:
\begin{equation}
    \mathcal{L}_{\text{m}} =  \frac{1}{N} 
    \sum_{\delta,\theta} (||\Delta x_{\delta,\theta}^-||_2 + ||\Delta x_{\delta,\theta}^+||_2).
\end{equation}
The final objective combines all terms:
\begin{equation}
    \mathcal{L}_{\text{total}} = \mathcal{L}_{\text{rt}} + 
    \lambda_{\text{oc}} \mathcal{L}_{\text{oc}} + 
    \lambda_{\text{p}} \mathcal{L}_{\text{p}} +
    \lambda_{\text{m}} \mathcal{L}_{\text{m}},
\end{equation}
where $\lambda{\text{oc}}$, $\lambda_{\text{p}}$, and $\lambda_{\text{m}}$ are weighting factors that balance temporal regularization, sparsity, and motion constraints.

\begin{figure}[t]
    \centering
    \includegraphics[width=\linewidth]{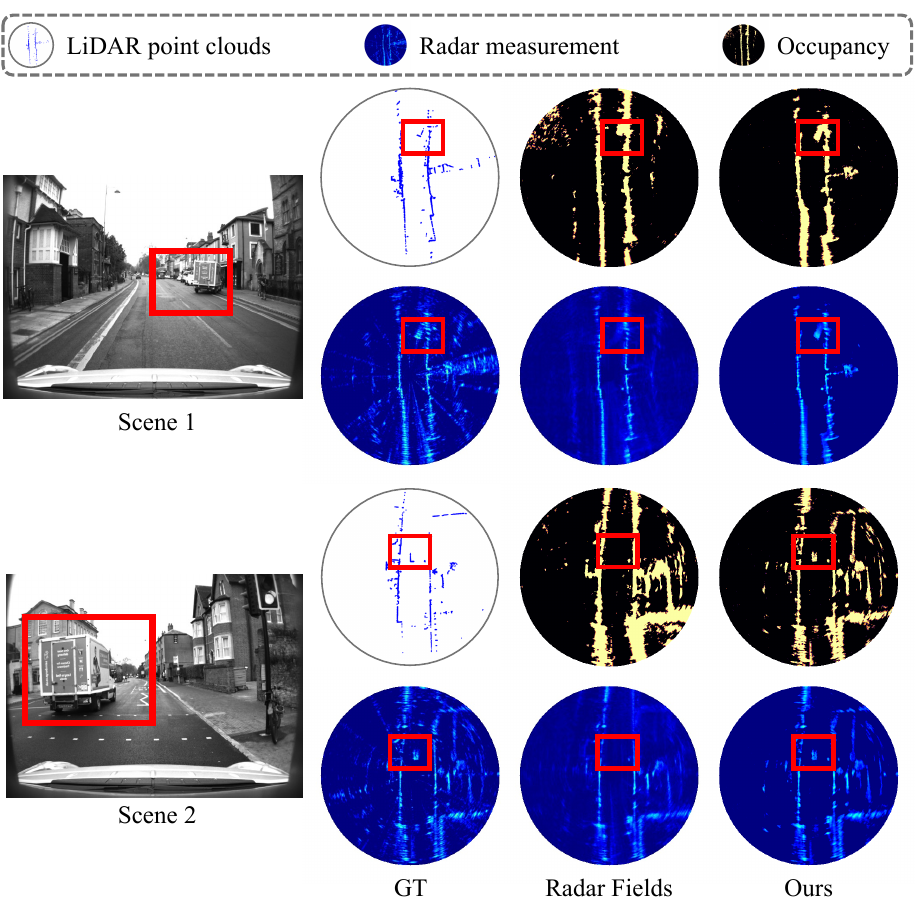}
    \caption{Qualitative comparison of novel-view radar measurement synthesis and occupancy estimation on the Oxford Radar RobotCar dataset~\cite{RadarRobotCarDatasetICRA2020}. Ground-truth occupancy is derived from LiDAR point clouds. RF4D reconstructs radar measurements with clear structures and preserved dynamic targets (red boxes), while Radar Fields produces noisier and blurrier results.}
    \label{fig:fig3}
\end{figure}
\begin{table*}[t]
\centering
\scriptsize
\setlength{\tabcolsep}{2pt}
\caption{Comparison of occupancy estimation results across different driving scenarios on the Oxford Radar RobotCar~\cite{RadarRobotCarDatasetICRA2020} and Boreas~\cite{burnett_ijrr23} datasets.
The best and second-best results are highlighted in \textbf{bold} and \underline{underline}, respectively.}
\begin{tabular}{c|cc|cc|cc|cc|cc|cc|cc}
\toprule
& \multicolumn{8}{c|}{\textbf{Oxford RobotCar}} 
&\multicolumn{6}{c}{\textbf{Boreas}} \\

 & \multicolumn{2}{c}{\textbf{Scene 1}} 
  & \multicolumn{2}{c}{\textbf{Scene 2}} 
  & \multicolumn{2}{c}{\textbf{Scene 3}}
  & \multicolumn{2}{c|}{\textbf{Scene 4}}
  & \multicolumn{2}{c}{\textbf{Sun}}
  & \multicolumn{2}{c}{\textbf{Rain}}
  & \multicolumn{2}{c}{\textbf{Static}}\\
\textbf{Method}  & CD\textdownarrow & RCD\textdownarrow 
                & CD\textdownarrow & RCD\textdownarrow 
                & CD\textdownarrow & RCD\textdownarrow
                & CD\textdownarrow & RCD\textdownarrow
                & CD\textdownarrow & RCD\textdownarrow
                & CD\textdownarrow & RCD\textdownarrow
                & CD\textdownarrow & RCD\textdownarrow\\
\midrule
CFAR~\cite{richards2005fundamentals}

  & 19.5299 & 0.0276
  & 30.5561 & 0.0387
  & 7.9197 & \underline{0.0185}
  & 26.4644 & 0.0291
  & 8.7570 & 0.0214
  & \underline{8.7636} & 0.0234
  & 13.7390 & 0.0083 \\

Bayesian filtering~\cite{werber2015automotive}

  & \underline{13.2786} & \textbf{0.0179}
 &  \underline{21.7657} & \underline{0.0273}
  &  \underline{6.3696}& 0.0208
  &  \textbf{18.6878}& \underline{0.0206} 
  &  \underline{6.0716}& \underline{0.0150} 
  &  \textbf{8.6918}& \underline{0.0069} 
  &  \underline{8.4866}& \underline{0.0049}  \\
\midrule
Ours
& \textbf{7.3449} & \underline{0.0190}
& \textbf{8.0909} & \textbf{0.0125}
& \textbf{3.2896} & \textbf{0.0097}
& \underline{18.7607} & \textbf{0.0147}
& \textbf{5.4482} & \textbf{0.0033}
 & 10.4558 & \textbf{0.0051} 
  & \textbf{8.0739} & \textbf{0.0040}
  \\

\bottomrule
\end{tabular}
\label{tab:occupancy}
\end{table*}

\section{Experiments}
\subsection{Experimental setup}
\paragraph{Datasets.}  
We evaluate RF4D on two publicly available radar datasets: Oxford Radar RobotCar~\cite{RadarRobotCarDatasetICRA2020} and Boreas~\cite{burnett_ijrr23}.
Both datasets contain spinning FMCW radar measurements capturing full 360-degree range-azimuth scans, synchronized with auxiliary sensors including GPS/IMU, cameras, and LiDAR.
Accurate vehicle odometry is provided, enabling precise radar pose estimation for each frame. 
The datasets cover diverse driving scenarios with moving vehicles, making them suitable for evaluating radar-based novel view synthesis in dynamic outdoor environments.
For the Boreas dataset, we select four representative scenes: \textit{sun}, \textit{snow}, \textit{rain}, and \textit{static}.
The \textit{sun}, \textit{snow}, and \textit{rain} scenes contain dynamic traffic, while static has no moving objects.
The \textit{snow} and \textit{rain} scenes are used to assess performance under adverse weather conditions.
For the Oxford Radar RobotCar dataset, which only includes clear-weather data, we randomly choose four scenes.
Each scene consists of approximately 60–100 consecutive radar scans, among which four scans are uniformly sampled for testing, and the remaining scans are used for training.
\paragraph{Data preprocessing.}  
Following the preprocessing strategy outlined in RadarFields~\cite{borts2024radar}, we discard the innermost 50 range bins (approximately a 3-meter radius around the radar sensor) from all radar measurements.
These bins predominantly reflect the metallic rooftop of the data collection vehicle and thus do not contribute meaningful scene information.
We retain the radar measurements up to the maximum range bin of 1200 (approximately 70 meters), resulting in 1150 effective range bins.
All azimuth angles (totaling 400) are retained without modification, yielding a final size of $(400, 1150)$ per radar measurement.
Additionally, each scene's coordinates are uniformly scaled to fit within a normalized cubic volume $[-1, 1]^3$, and timestamps are similarly normalized to the range $[0, 1]$.

\paragraph{Implementation details.}  
Our RF4D model is implemented in PyTorch~\cite{paszke2019pytorch} and trained on a single NVIDIA RTX A5000 GPU.
For efficient neural field optimization, we use the tiny-cuda-nn framework~\cite{tiny-cuda-nn} to implement neural network backbones and input encodings.
The model is optimized using the Adam optimizer~\cite{kingma2014adam} with an initial learning rate of $1\times 10^{-4}$, which is progressively decayed by a factor of 0.1 during training, resulting in a final learning rate of $1\times 10^{-5}$.
Each scene is trained independently for 15000 iterations. In each training iteration, we randomly sample a batch of 4 radar scans from the input sequence and further randomly select range-azimuth bins from these frames for supervision.
\paragraph{Evaluation metrics.}
In line with prior work~\cite{borts2024radar}, we evaluate novel radar measurement synthesis with Peak Signal to Noise Ratio (PSNR).
We additionally include the Structural Similarity Index (SSIM) to assess perceptual and structural fidelity, which complements PSNR by being more sensitive to local contrast and structural patterns, including the shapes and boundaries of dynamic targets.
For occupancy estimation, we report Chamfer Distance (CD) and Relative Chamfer Distance (RCD) between predicted 2D bird’s eye view point clouds and LiDAR derived references.
Since accurate ground-truth occupancy is difficult to extract directly from radar due to low azimuth resolution, multipath interference, and inherent sensor noise, we leverage synchronized LiDAR point clouds to construct geometric occupancy references.
For each radar frame, we calibrate the corresponding LiDAR point clouds to radar coordinates using the provided LiDAR-to-radar transformation matrix. We then select points within the radar's sensing range bounds and with $-1 < z < 1$. The resulting points are projected onto the $xy$-plane to generate the BEV ground-truth occupancy for that frame.
To ensure compatibility with dynamic scenes, we perform occupancy evaluation using only the LiDAR point clouds corresponding to each radar timestamp, without accumulating across time.
\subsection{Comparison with state-of-the-art}
To ensure fair and rigorous evaluation, we focus exclusively on NF-based approaches, excluding methods relying on handcrafted environment modeling or sensor-specific signal simulations.
We compare RF4D with three representative state-of-the-art methods: Radar Fields~\cite{borts2024radar}, D-NeRF~\cite{pumarola2021d}, and Hexplane~\cite{cao2023hexplane}.
For Radar Fields, we directly adopt their official implementation, ensuring consistency and fair comparison.
For D-NeRF and Hexplane, originally designed for RGB-based dynamic NVS, we adapt their architectures to our radar-based pipeline and retrain accordingly.
\paragraph{Quantitative comparison.}
Tables~\ref{tab:oxford_results} and~\ref{tab:boreas_results} present quantitative comparisons on the Oxford Radar RobotCar and Boreas datasets, respectively.
RF4D consistently achieves the best performance across most driving scenarios on both datasets, outperforming all competing methods in both radar measurement synthesis (PSNR, SSIM) and occupancy estimation (CD, RCD).
The jointly superior PSNR and SSIM results show that RF4D achieves both accurate and structurally consistent radar synthesis.
Meanwhile, the lower CD and RCD values show that the predicted occupancies are spatially more precise and better aligned with LiDAR-derived geometry.
In contrast, methods like D-NeRF and HexPlane, designed for RGB data without radar-specific modeling, yield inferior radar synthesis and occupancy estimation.
Although Radar Fields incorporates radar sensing physics, its Bayesian filtering occupancy supervision is prone to multipath-induced false positives, which limit overall performance.

\begin{figure}[t]
    \centering
    \includegraphics[width=\linewidth]{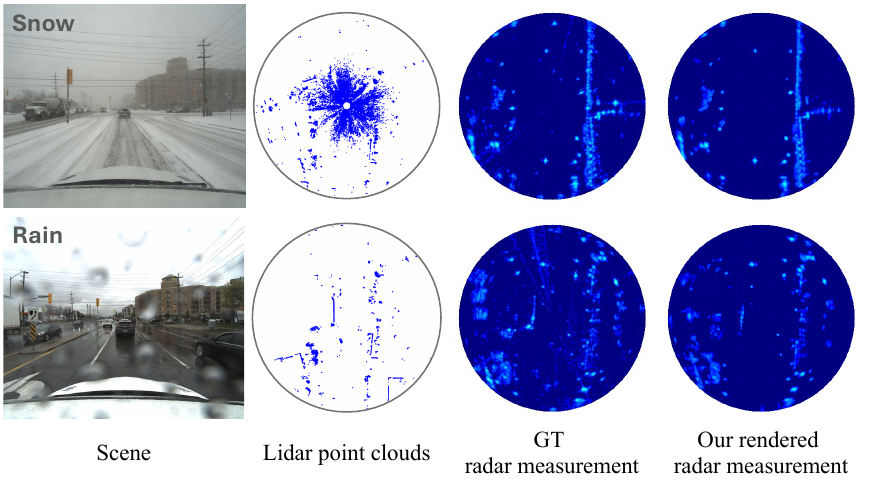}
    \caption{Radar robustness and generalization across weather conditions. While LiDAR point clouds degrade severely in snow, radar measurements remain stable. RF4D accurately reconstructs radar measurements, maintaining consistent performance across different weather conditions.}
    \label{fig:fig4}
\end{figure}
\paragraph{Qualitative comparison.}
Figure~\ref{fig:fig3} presents qualitative comparisons of reconstructed radar measurements and occupancy estimation results across two representative scenes using Radar Fields~\cite{borts2024radar} and our method.
Our model successfully recovers dynamic objects (highlighted in red boxes), demonstrating its ability to capture temporal scene variations.
We attribute this improvement to the incorporation of temporal information and temporal regularization, which enable RF4D to maintain consistent estimation of dynamic targets across frames.
In addition, RF4D renders significantly cleaner radar reconstructions and predicts more accurate occupancy estimations that closely align with the LiDAR-derived ground truth.
Radar measurements are often affected by sensor noise and multipath interference, which introduce false-positive responses.
Radar Fields tends to misinterpret these spurious reflections as occupied regions, leading to false and unstable occupancy predictions.
In contrast, RF4D employs a Gumbel–Sigmoid activation coupled with an occupancy gating mechanism during radar-specific power rendering, driving the neural field toward confident, near-binary occupancy outputs and effectively suppressing uncertain responses.
Combined with the global occupancy sparsity regularization, these designs yield cleaner rendered radar measurements and more reliable occupancy maps.
Figure~\ref{fig:fig4} further verifies the radar robustness and generalization across weather conditions.
While LiDAR point clouds become noisy and incomplete in snow conditions, radar measurements remain stable and reliable.
Our RF4D framework accurately reproduces these radar measurements, showing consistent performance across different weather conditions.

\subsection{Ablation study}
In Table~\ref{tab:oxford_ablation} and~\ref{tab:boreas_ablation}, we conduct ablation studies to analyze the contribution of each loss component.
When trained with only the radar power reconstruction loss $\mathcal{L}_{rt}$, the model achieves acceptable PSNR but performs poorly in SSIM and occupancy metrics (CD and RCD).
This is because, without occupancy constraints, the network tends to predict all regions as occupied ($\alpha \approx 1$) and overfit the radar power responses everywhere, resulting in inaccurate and structureless occupancy predictions.
Introducing the occupancy regularization loss $\mathcal{L}_{p}$ encourages the model to focus on confident and stable regions, effectively suppressing uncertain occupancy estimates and improving both measurement synthesis and geometric accuracy.
Note that when removing the temporal regularization losses $\mathcal{L}_{oc}$ and $\mathcal{L}_{m}$, we also remove the scene flow module $f_{\Delta x}$, since these components are designed to work together.
Adding the temporal consistency loss $\mathcal{L}_{oc}$ further enforces temporal alignment of occupancy across frames by warping each point to its adjacent frames using the predicted motion offsets.
However, this alone does not yield direct improvement, as the motion offsets can become unstable without explicit constraints.
After further introducing the motion regularization term $\mathcal{L}_{m}$ to constrain the magnitude and smoothness of these offsets, the temporal predictions become more coherent, leading to consistent gains across all metrics.

\subsection{Occupancy estimation comparison}
In this section, we further compare our method with traditional radar occupancy estimation approaches, including Constant False Alarm Rate (CFAR)~\cite{richards2005fundamentals} and Bayesian filtering~\cite{werber2015automotive}.
As shown in Table~\ref{tab:occupancy}, our method consistently outperforms other approaches across most scenes.
Conventional methods such as CFAR and Bayesian filtering often preserve multipath and noise artifacts, leading to unstable and inaccurate occupancy estimation.
In contrast, our method provides accurate and stable occupancy estimates.
This indicates that using occupancy maps from traditional methods as supervision during training may introduce misleading signals and hinder learning of precise scene structure.
Notably, our approach does not rely on any external occupancy estimator for supervision, but learns directly from radar measurements.

\begin{figure}
    \centering
    \includegraphics[width=\linewidth]{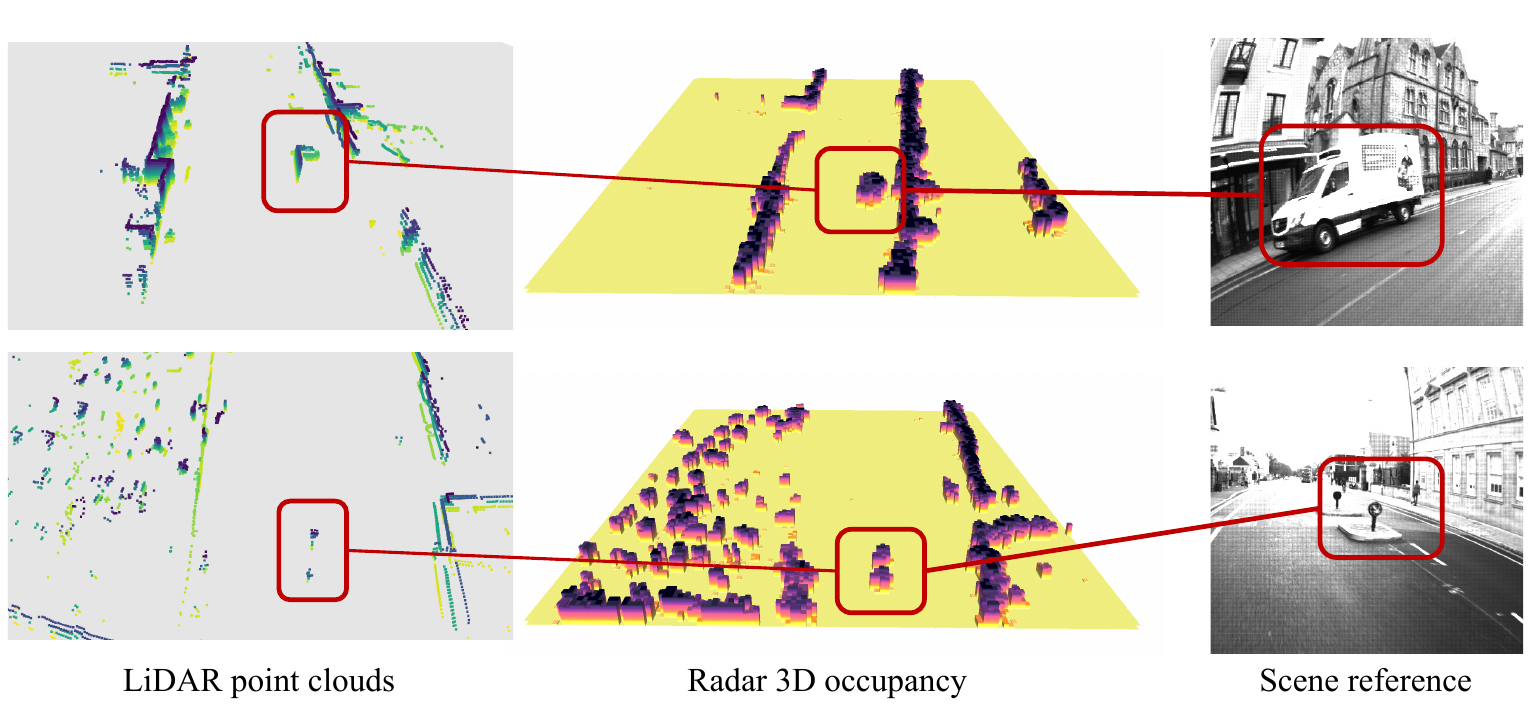}
    \caption{3D voxel grid reconstruction from radar measurements. Our method reconstructs full 3D occupancy geometry from sparse and low-resolution radar data, capturing both moving vehicles and static objects present in the scene.}
    \label{fig:3d occupancy}
\end{figure}
\subsection{3D voxel grid reconstruction}
We adopt physics-based importance sampling~\cite{borts2024radar}, which uses radar antenna gain profiles to obtain elevation information to further reconstruct 3D geometry. As shown in Figure~\ref{fig:3d occupancy}, our RF4D with physics-based importance sampling can reconstruct full 3D occupancy geometry of the scene from sparse and low-resolution radar data, successfully capturing both moving vehicles and static objects in the scene.
However, it is important to note that the elevation field of view (FoV) of the radar is limited to around 2 degrees, which restricts the reconstruction ability for objects located outside the radar's elevation coverage.

\section{Conclusion}
In this work, we introduced RF4D, a radar-based neural field framework for novel view synthesis in dynamic outdoor scenes.
By integrating temporal modeling with a dedicated scene flow module, RF4D enforces temporal coherence and accurately captures moving targets.
The proposed radar-specific power rendering further aligns the neural field representation with radar sensing physics, enhancing both accuracy and interpretability.
Comprehensive experiments on public datasets demonstrate that RF4D substantially outperforms existing methods in radar measurement synthesis and occupancy estimation, particularly under dynamic and adverse conditions.
These results highlight the potential of radar-based neural fields for robust 4D scene understanding.

\section*{Acknowledgment}
This work was carried out at the Rapid-Rich Object Search (ROSE) Lab, School of Electrical \& Electronic Engineering, Nanyang Technological University (NTU), Singapore.
{
    \small
    \bibliographystyle{unsrt}
    \bibliography{main}
}



\end{document}